\documentclass[letterpaper]{article} % DO NOT CHANGE THIS
\usepackage{aaai2026}  % DO NOT CHANGE THIS
\usepackage{times}  % DO NOT CHANGE THIS
\usepackage{helvet}  % DO NOT CHANGE THIS
\usepackage{courier}  % DO NOT CHANGE THIS
\usepackage[hyphens]{url}  % DO NOT CHANGE THIS
\usepackage{graphicx} % DO NOT CHANGE THIS
\usepackage{natbib}  % DO NOT CHANGE THIS AND DO NOT ADD ANY OPTIONS TO IT
\usepackage{caption} % DO NOT CHANGE THIS AND DO NOT ADD ANY OPTIONS TO IT
\usepackage{algorithm}
\usepackage{algorithmic}
\usepackage{multirow}
\usepackage{booktabs}
\usepackage{amsmath}

\usepackage{newfloat}
\usepackage{listings}
\DeclareCaptionStyle{ruled}{labelfont=normalfont,labelsep=colon,strut=off} % DO NOT CHANGE THIS
\floatstyle{ruled}
\newfloat{listing}{tb}{lst}{}
\floatname{listing}{Listing}
\usepackage{bm}

\def\vp{{\mathbf{p}}}

\def\vq{{\mathbf{q}}}
\def\vo{{\mathbf{o}}}
\def\vc{{\mathbf{c}}}
\def\va{{\mathbf{a}}}

\def\vV{{\mathbf{V}}}

\def\vpi{{\bm{\pi}}}

\title{Efficient Reasoning for Large Reasoning Language Models via Certainty-Guided Reflection Suppression}

\author {
Jiameng Huang\textsuperscript{\rm 1}\thanks{These authors contributed equally.}, Baijiong Lin\textsuperscript{\rm 2}\footnotemark[1], Guhao Feng\textsuperscript{\rm 1}, Jierun Chen\textsuperscript{\rm 3}, Di He\textsuperscript{\rm 1}\thanks{Corresponding to Lu Hou \texttt{<houlu3@huawei.com>} and Di He \texttt{<dihe@pku.edu.cn>}.}, Lu Hou\textsuperscript{\rm 3}\footnotemark[2]
}

\affiliations {
    \textsuperscript{\rm 1} National Key Laboratory of General Artificial Intelligence,\\ School of Intelligence Science and Technology, Peking University\\
    \textsuperscript{\rm 2} The Hong Kong University of Science and Technology (Guangzhou)\\
    \textsuperscript{\rm 3} Huawei Technologies Co., Ltd. \\
    \{huangjm, fenguhao\}@stu.pku.edu.cn, \{bj.lin.email, jierunchen\}@gmail.com, dihe@pku.edu.cn,  houlu3@huawei.com
}
\usepackage{bibentry}
\begin{document}

\maketitle

% ============= 0 ABSTRACT ================

\begin{abstract}

Recent Large Reasoning Language Models (LRLMs) employ long chain-of-thought reasoning with complex reflection behaviors, typically signaled by specific trigger words (e.g., ``Wait" and ``Alternatively") to enhance performance. However, these reflection behaviors can lead to the overthinking problem where the generation of redundant reasoning steps that unnecessarily increase token usage, raise inference costs, and reduce practical utility. In this paper, we propose \textbf{Certainty-Guided Reflection Suppression (CGRS)}, a novel method that mitigates overthinking in LRLMs while maintaining reasoning accuracy. CGRS operates by dynamically suppressing the model's generation of reflection triggers when it exhibits high confidence in its current response, thereby preventing redundant reflection cycles without compromising output quality. Our approach is model-agnostic, requires no retraining or architectural modifications, and can be integrated seamlessly with existing autoregressive generation pipelines.
Extensive experiments across four reasoning benchmarks (i.e., AIME24, AMC23, MATH500, and GPQA-D) demonstrate CGRS's effectiveness: it reduces token usage by an average of $18.5\%$ to $41.9\%$ while preserving accuracy and also achieves the optimal balance between length reduction and performance compared to state-of-the-art baselines. These results hold consistently across model architectures (e.g., DeepSeek-R1-Distill series, QwQ-32B, and Qwen3 family) and scales (4B to 32B parameters), highlighting CGRS's practical value for efficient reasoning. 

\end{abstract}

% ============= 0 ABSTRACT ================

% ============= 1 INTRO ================

\begin{figure}[t!]
 \centering
 \includegraphics[width=0.48\textwidth]{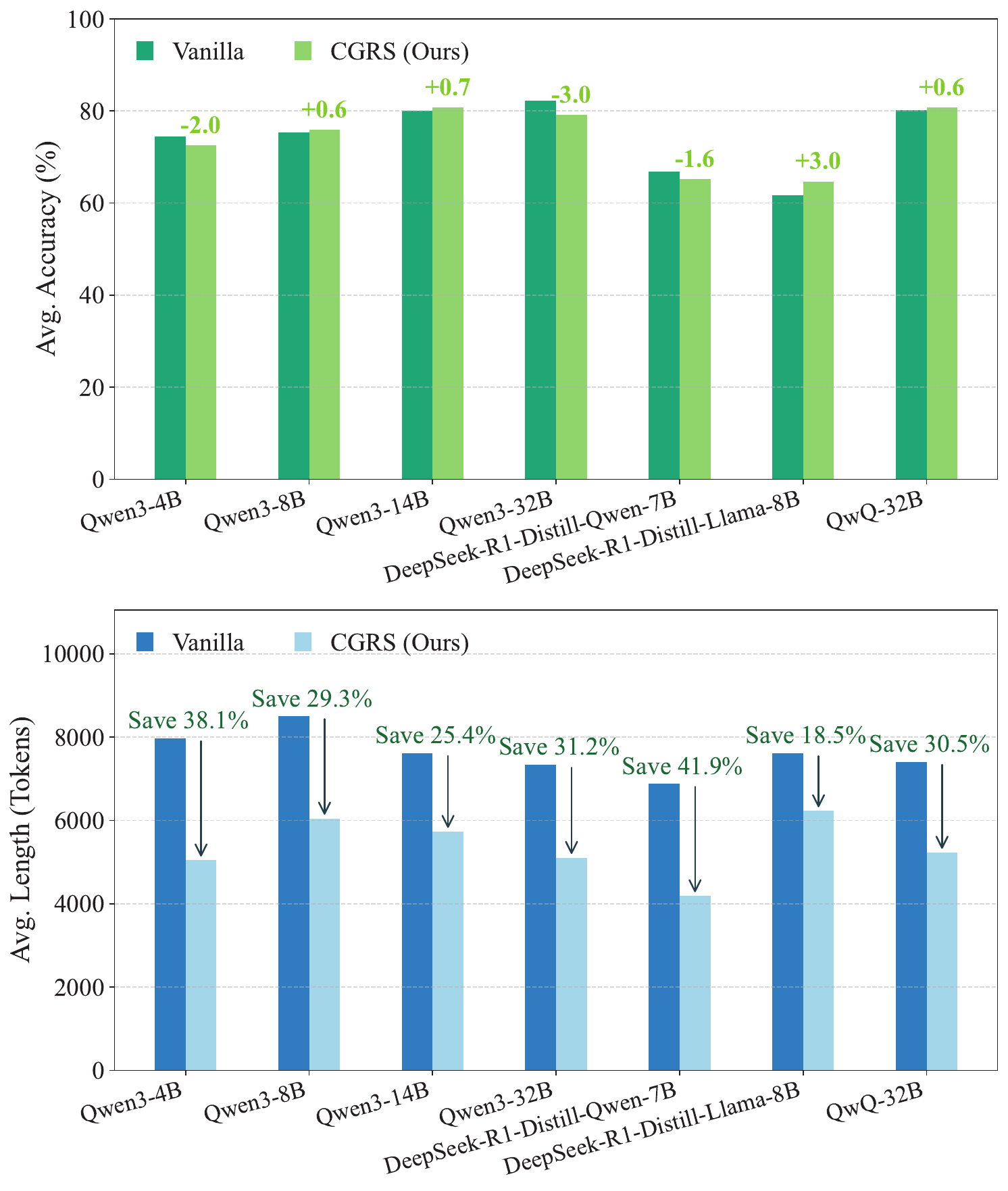}
 \caption{Accuracy and output length of the Vanilla and proposed CGRS methods across different models, averaged over three mathematical reasoning benchmarks (AIME24, AMC23, and MATH500) and one scientific reasoning benchmark (GPQA-D). CGRS achieves an average token reduction of $18.5-41.9\%$ while maintaining performance.
 }
 \label{fig:main}
\end{figure}

\section{Introduction}

Large Reasoning Language Models (LRLMs), including OpenAI’s o1/o3~\citep{o1,o3} and DeepSeek-R1~\citep{guo2025deepseekr1}, have demonstrated remarkable performance on demanding benchmarks, especially in advanced mathematics and program synthesis.
A key factor behind their success is their slow-thinking capability, which combines step-by-step deduction with complex reflection behaviors, including backtracking, exploring alternative strategies, and self-verification of results ~\citep{zhu2025conciseadaptivethinkinglarge}. 

Although these reflection behaviors enable models to self-correct and obtain accurate answers, they often suffer from a significant \textbf{overthinking} problem~\citep{cuadron2025dangeroverthinkingexaminingreasoningaction, sui2025stopoverthinkingsurveyefficient, fan2025missingpremiseexacerbatesoverthinking}, where LRLMs persistently continue reasoning even after arriving at correct solutions. This phenomenon leads to substantial increases in token consumption, higher inference costs, and degraded user experience due to unnecessary delays~\citep{sun2025invisibletokensvisiblebills, wang2025harnessingreasoningeconomysurvey}. 
In extreme cases, excessively prolonged responses may exceed context window limits, resulting in truncated critical information and compromised accuracy~\citep{li2025thinklesstrainingfreeinferenceefficientmethod}. 
Addressing this overthinking problem and developing efficient reasoning mechanisms, therefore represents a crucial challenge for improving LRLM performance and practicality.

To address this challenge, training-free methods have recently emerged as lightweight solutions that mitigate overthinking without requiring additional training \cite{zhu2025conciseadaptivethinkinglarge}. 
For instance, prompt-guided approaches like TALE \cite{han-etal-2025-token} prompts model responses within predefined token budgets through carefully designed instructions. 
While straightforward, these methods rely heavily on the model's innate instruction-following capabilities. 
Decoding-manipulation techniques, such as Dynasor \cite{fu2025efficientlyscalingllmreasoning} and DEER \cite{yang2025dynamicearlyexitreasoning}, dynamically adjust the decoding process to eliminate redundancy by terminating generation when the model reaches sufficient confidence. However, their effectiveness remains highly sensitive to the design of early-exit conditions.
Consequently, current methods lack adaptive mechanisms to properly balance reflection suppression with reasoning quality, often either excessively suppressing valid corrections or inadequately curbing redundant reflections.

We propose Certainty-Guided Reflection Suppression (\textbf{CGRS}), a novel method that mitigates overthinking in LRLMs while preserving reasoning accuracy.
Previous work has demonstrated that reflection behaviors are typically signaled by specific \textbf{reflection triggers}, i.e., keywords like ``Wait", ``Alternatively", ``But", and ``Hmm" \citep{muennighoff2025s1,zhang2025alphaonereasoningmodelsthinking, ding2025thinkingtokenshelptrap}. Our CGRS approach proactively suppresses the model's generation of these reflection triggers when it exhibits high certainty in its current response, thereby preventing redundant reflection cycles and effectively alleviating overthinking while preserving reasoning quality.

Specifically, CGRS operates through two key components: 
(i) \textbf{Certainty estimation}: We first identify logical breakpoints in the reasoning process using structural delimiters (e.g., \verb|\n\n|), then probe tentative final answers by injecting the prompt \texttt{**Final Answer: \textbackslash boxed}. 
The model's certainty in its current response is quantified as the average token entropy over these tentative answers. 
Note that a lower entropy indicates higher confidence, suggesting a diminished need for further reflection. 
(ii) \textbf{Suppression of dynamic reflection triggers}: Based on the certainty score, we probabilistically suppress reflection triggers (e.g. ``Wait", ``But") by setting their logits to large negative values during sampling, effectively preventing unnecessary reflection cycles while preserving valid corrective behaviors.

We evaluate CGRS across diverse open-source LRLMs spanning multiple architectures and scales: the \textit{DeepSeek-R1-Distill} series (Qwen-7B and Llama-8B)~\citep{guo2025deepseekr1}, \textit{QwQ-32B}~\citep{qwq32b}, and the \textit{Qwen3} family (4B, 8B, 14B, and 32B)~\citep{qwen3}. Our evaluation covers four reasoning benchmarks, including mathematical tasks (AIME24 \cite{aime}, AMC23 \citep{amc23}, and MATH500 \cite{math500}) and scientific reasoning (GPQA Diamond \citep{gpqa}).
As shown in Fig.~\ref{fig:main}, CGRS achieves significant token reduction ranging from $18.5\%$ to $41.9\%$, while maintaining comparable accuracy.
Furthermore, CGRS outperforms existing efficient reasoning approaches, including both prompt-guided and decoding-manipulation methods, by delivering greater token efficiency with minimal accuracy impact.
These results solidify CGRS as a superior solution for enabling efficient reasoning in LRLMs.

Our contributions are summarized as follows:
\begin{itemize}
	\item We propose CGRS, a training-free efficient reasoning method that alleviates the overthinking problem in LRLMs by dynamically suppressing reflection behaviors based on the model's internal certainty signals;
	\item We introduce two key components: (i) a {certainty estimation} mechanism that quantifies model confidence through entropy analysis of tentative answers, and (ii) a {dynamic reflection trigger suppression} technique that selectively disables unproductive reflection cycles; 
	\item We validate our method against state-of-the-art efficient reasoning methods
	spanning multiple model scales (4B to 32B parameters) on four reasoning benchmarks. 
	Across all evaluations, CGRS demonstrates superior token efficiency with minimal impact on accuracy, positioning it as a superior solution for efficient reasoning in LRLMs.
\end{itemize}

% ============= 1 INTRO ================

% ============= 2 RELATED WORK ================

\section{Related Work}

LRLMs have significantly advanced long Chain-of-Thought (CoT) reasoning through supervised fine-tuning and reinforcement learning \cite{Pan2025ASO}. 
However, prolonged CoT often introduces redundant steps that increase computational costs, a phenomenon known as \textbf{overthinking} \cite{cuadron2025dangeroverthinkingexaminingreasoningaction, sui2025stopoverthinkingsurveyefficient, fan2025missingpremiseexacerbatesoverthinking}. 
To address this issue, many works focus on compressing verbose CoTs into more concise reasoning traces \cite{sui2025stopoverthinkingsurveyefficient, qu2025surveyefficientreasoninglarge, feng2025efficientreasoningmodelssurvey, wang2025harnessingreasoningeconomysurvey, zhu2025conciseadaptivethinkinglarge}, which can be broadly categorized into two approaches: training-based methods \cite{arora2025traininglanguagemodelsreason, aggarwal2025l1controllinglongreasoning,wang2025learningthinkinformationtheoreticreinforcement,ma2025cotvalvelengthcompressiblechainofthoughttuning, munkhbat2025selftrainingelicitsconcisereasoning, chen2025reasoninglanguagecomprehensivesurvey, zhu2025surveylatentreasoning} and training-free methods \cite{ma2025reasoningmodelseffectivethinking,yu2025premisescalablestrategicprompt,han-etal-2025-token, yang2025speculativethinkingenhancingsmallmodel, kimiteam2025kimik15scalingreinforcement, wu2025unlockingefficientlongtoshortllm, fu2025efficientlyscalingllmreasoning,jiang2025flashthinkearlyexitmethod,yang2025dynamicearlyexitreasoning,li2025steeringllmthinkingbudget, aytes2025sketchofthoughtefficientllmreasoning, yang2025speculativethinkingenhancingsmallmodel}. 
While effective, training-based methods often require retraining, careful data curation, and may compromise the general capabilities of the pretrained model. 
In this work, we focus on training-free methods, which eliminate the need for additional training and can be further divided into four categories: prompt-guided, pipeline-based, model-merging, and decoding-manipulation methods.

\paragraph{Prompt-guided Methods.} This type of method \cite{ma2025reasoningmodelseffectivethinking,yu2025premisescalablestrategicprompt,han-etal-2025-token} compress lengthy CoT reasoning through prompt engineering. 
For instance, TALE \cite{han-etal-2025-token} simply prompts the model to solve the problem within token budgets. 
While simple, these methods heavily depend on the model's inherent instruction-following capability.

\paragraph{Pipeline-based Methods.} This category of methods \cite{ong2025routellmlearningroutellms, aytes2025sketchofthoughtefficientllmreasoning, yang2025speculativethinkingenhancingsmallmodel} distribute queries or reasoning stages across multiple LRLMs, leveraging the efficiency of small models and the advanced reasoning of larger ones. 
However, these approaches introduce deployment complexity and additional inference overhead from auxiliary modules.

\paragraph{Model Merging.} Such methods \cite{kimiteam2025kimik15scalingreinforcement, wu2025unlockingefficientlongtoshortllm} combine slow- and fast-thinking models via weight averaging to enable long-to-short reasoning. 
While effective for medium-sized models, this approach underperforms at extreme scales (very large or small models) and lacks precise control over reasoning depth \cite{zhu2025conciseadaptivethinkinglarge}.

\paragraph{Decoding-manipulation Methods.}  This type of method \citep{fu2025efficientlyscalingllmreasoning,jiang2025flashthinkearlyexitmethod,yang2025dynamicearlyexitreasoning,li2025steeringllmthinkingbudget} dynamically adjusts the decoding process to eliminate redundancy. For example, Dynasor \cite{fu2025efficientlyscalingllmreasoning} and DEER \cite{yang2025dynamicearlyexitreasoning} terminate decoding early when the model reaches sufficient confidence, though their performance depends critically on the exit condition design.

In this work, we focus on decoding manipulation for efficient reasoning. Our proposed CGRS method dynamically intervenes in token logits to suppress reflective behaviors based on the model's internal certainty signals.

% ============= 2 RELATED WORK ================

% ============= 3 METHOD ================

\section{Method}
\label{sec:method}

% \footnote{\hou{try merge the fig and description of the case study from the expt sec to here, for a clearer illustration of  our motivation and our proposed method. }}
In this section, we present our Certainty-Guided Reflection Suppression (\textbf{CGRS}) method for efficient reasoning in LRLMs.

\subsection{Preliminaries}

Large reasoning language models demonstrate unique generation patterns during inference. Given a problem $\vq$ (with prompt included), the output $\vo$ is generated token by token as $\vo_t \sim \vpi(\cdot|\vq,\vo_{<t})$, where $\vpi$ is the model. 
LRLMs organize their output using thinking delimiters (i.e., \texttt{<think>} and \texttt{</think>}), splitting the response into two key parts: slow thinking and conclusion. 
During slow thinking, LRLMs perform detailed, step-by-step reasoning before summarizing their thought process and delivering the final answer in the conclusion~\citep{zhang2025alphaonereasoningmodelsthinking}.

Within slow thinking, models engage in complex reflection behaviors, including backtracking, exploring alternative approaches, and verifying results. 
These behaviors are often signaled by specific reflection triggers, i.e., keywords like ``Wait'', ``Alternatively'', ``But'', and ``Hmm''~\citep{muennighoff2025s1,zhang2025alphaonereasoningmodelsthinking, ding2025thinkingtokenshelptrap}.

Notably, these reflection behaviors cause models to continue reasoning even after reaching an initial answer. 
They may perform extra validation steps or switch reasoning strategies in later stages, either to double-check previous results or explore different paths. 
However, this can sometimes lead to unproductive overthinking, where the model enters unnecessary reasoning loops, repeatedly validating or re-evaluating without meaningful progress~\citep{chen2025think23overthinkingo1like}.

To address this, our CGRS method proactively suppresses the model's tendency to generate reflection triggers when it exhibits high certainty in its current response. 
This prevents redundant reflection behavior and avoids overthinking.
CGRS operates in two phases: (i) certainty estimation through checkpoint probing, and (ii) dynamic trigger suppression. We detail these components as follows.

\subsection{Certainty Estimation}

To estimate the necessity of reflective behaviors during inference, we introduce the \textbf{certainty score} that quantifies the LLM's confidence in its currently generated response.
Specifically, we establish \textbf{checkpoints} along the LLM's reasoning path, identified by specific tokens. 
\verb|\n\n| is used as the checkpoint marker, as this structural delimiter naturally indicates a thinking breakpoint~\citep{yang2025speculativethinkingenhancingsmallmodel}. 
At each checkpoint, we probe tentative final answers by appending a prompt $\vc=\texttt{**Final Answer: \textbackslash boxed}$ to the current generation $\vo_{<t}$, then concurrently generating a response $\va$. 
This probing runs independently of the main decoding process, preserving the integrity of the primary reasoning trajectory. 
Then we calculate the certainty score $C$ on the probed result $\va$ as follows:
\begin{align}
C = 1 - \left( \frac{\frac{1}{n} \sum_{i=1}^{n} \mathcal{H}(\vp_{\va_i})}{\log(|\vV|)} \right),
\label{eq:certainty}
\end{align}
where $n=|\va|$ is the token length of the probed results, 
$\vp_{\va_i}=\vpi(\cdot|\vq,\vo_{<t}, \vc, \va_{<i})$ represents the probability distribution of $\va_i$, $\mathcal{H}(\vp)=-\vp^\top\log(\vp)$ denotes the token entropy, and $|\vV|$ is the total number of tokens in the LLM's vocabulary. The term $\log(|\vV|)$ represents the maximum possible entropy (i.e., when the probability distribution is uniform across all tokens), serving as a normalization factor. 
This normalization bounds the certainty score $C$ to $[0, 1]$, where a higher $C$ indicates higher LLM confidence (i.e., lower entropy in $\va$), suggesting a reduced need for subsequent reflection behaviors.

\begin{algorithm}[!t]
\caption{The token prediction process in CGRS.}
\label{alg:reflection_trigger_control}
\textbf{Require}: LRLM $\vpi$, input prompt $\vq$, generated tokens $\vo_{<t}$, checkpoint prompt $\vc$, checkpoint marker \verb|\n\n|, set of reflection triggers tokens $S_{\text{trigger}}$, previous suppression probability $p$ (set to $0$ if not exist), threshold $\delta$;
\begin{algorithmic}[1]
\STATE compute probability distribution $\vp_t = \vpi(\cdot|\vq, \vo_{<t})$;
\STATE sample $r \sim \mathrm{Bernoulli}(p)$;
\IF{$r=1$}
\STATE modify $\vp_t$: set logits of tokens in $S_{\text{trigger}}$ to a large negative value and re-normalize;
\ENDIF
\STATE sample the next token $\vo_t$ from $\vp_t$;
\IF{checkpoint marker appears}
\STATE obtain tentative answers $\va$ given the input $(\vq, \vo_{<t}, \vc)$; 
\STATE compute certainty score $C$ using $\va$ via Eq. (\ref{eq:certainty});
\STATE compute suppression probability $p$ using $C$ and $\delta$ via Eq. (\ref{eq:suppression_prob});
\ENDIF
\STATE \textbf{return}: the next token $\vo_t$, the suppression probability $p$.
\end{algorithmic}
\end{algorithm}

\subsection{Reflection Suppression}

After estimating the LLM's certainty regarding its current generation, the certainty score $C$ guides the suppression of subsequent reflection behaviors.

\paragraph{Reflection Triggers.} Following prior work \cite{yang2025speculativethinkingenhancingsmallmodel,fu2025efficientlyscalingllmreasoning,zhang2025alphaonereasoningmodelsthinking}, we consider three trigger categories:
(i) core hesitation and transition words (e.g., ``But" and ``Wait"),
(ii) alternative proposal markers (e.g., ``Alternatively"), and
(iii) colloquial contemplation cues (e.g., ``Hmm"),
all of which signal potential shifts in the LLM’s reasoning mode.
Since a single trigger word has multiple natural language variants (e.g., ``wait" as a lowercase variant of ``Wait") that map to different tokens in the vocabulary, we construct a comprehensive trigger token set $S_{\text{trigger}}$ through frequency analysis. Specifically, we identify all possible variants of each reflection trigger in the tokenizer's vocabulary and then analyze their generation frequencies in reasoning traces. This analysis was performed using reasoning traces from the \textit{DeepSeek-R1-Distill-Qwen-7B} model~\citep{guo2025deepseekr1} on both the AIME24~\citep{aime} and AMC23~\citep{amc23} benchmarks. To ensure robustness, we conducted four independent inference runs per dataset, aggregating statistics across all executions.
Additional details and the complete list of trigger words and their variants are available in Appendix.

\paragraph{Suppressing Triggers Generation via Certainty.} 
We reduce the reflection behaviors by suppressing the generation of triggers according to the certainty score. 
Specifically, we first calculate the suppression probability $p$ as follows,  
\begin{align}
p = \max\left(0, \frac{C - \delta}{1 - \delta}\right),
\label{eq:suppression_prob}
\end{align}
where $\delta\in[0,1]$ is the confidence threshold. 
Then, with probability $p$, we set the logits of trigger tokens in $S_{\text{trigger}}$ to a large negative value, thereby effectively excluding them from sampling. 
Eq. (\ref{eq:suppression_prob}) indicates that trigger suppression occurs only when $C>\delta$. Besides, higher $C$ values yield more frequent trigger suppression, as the LLM has exhibited high confidence in current responses, thereby requiring fewer subsequent reflection behaviors.

In summary, during inference, CGRS inserts checkpoints to probe intermediate answers for certainty calculation, then suppresses reflection tokens accordingly. 
This dynamic control mechanism reduces redundant reflection behaviors while preserving necessary corrections, enabling efficient reasoning. 
Moreover, CGRS is model-agnostic, requires no retraining or architectural modifications, and can be seamlessly integrated into existing autoregressive generation pipelines.
The token prediction process of CGRS is shown in Algorithm \ref{alg:reflection_trigger_control}.

% ============= 3 METHOD ================

% ============= 4 EXPERIMENT ================

% ==================== EXPERIMENTS ==========================
\section{Experiments}
\label{sec:experiments}

\begin{table*}[!t]
    \centering
    % \small
    % \footnotesize
    \resizebox{\linewidth}{!}{
    \begin{tabular}{l ccc ccc ccc ccc|cc} 
    \toprule
    \multirow{2.5}{*}{\textbf{Method}} & \multicolumn{3}{c}{\textbf{AIME24}} & \multicolumn{3}{c}{\textbf{AMC23}} & \multicolumn{3}{c}{\textbf{MATH500}} & \multicolumn{3}{c}{\textbf{GPQA-D}}
    & \multicolumn{2}{c}{\textbf{AVG}} \\
    \cmidrule(lr){2-4} \cmidrule(lr){5-7} \cmidrule(lr){8-10} \cmidrule(lr){11-13} \cmidrule(l){14-15}
     & Acc$\uparrow$ & Len$\downarrow$ & LR $\uparrow$ & Acc$\uparrow$ & Len$\downarrow$ & LR $\uparrow$ & Acc$\uparrow$ & Len$\downarrow$ & LR $\uparrow$ & Acc$\uparrow$ & Len$\downarrow$ & LR $\uparrow$ & Acc$\uparrow$ & LR $\uparrow$\\
    \midrule
    \multicolumn{15}{c}{\textit{\textbf{Qwen3-4B}}} \\
    Vanilla & 60.0 & 11449 & \text{-} & 91.7 & 7449 &  \text{-} & 92.7 & 4796 & \text{-} & 53.5 & 8195 & \text{-} & 74.5 & \text{-} \\
    TALE & 48.9 & 9727 & \text{15.0\%} & 86.7 & 5107 & \text{31.4\%} & 89.1 & 2657 & \text{44.6\%} & 36.2 & 4938 & \text{39.7\%} & 65.2 & \text{32.7\%}\\
    NoThinking & 24.4 & 4504 & \text{60.7\%} & 70.0 & 1710 & \text{77.0\%} & 84.9 & 988 & \text{79.4\%} & 47.5 & 1471 & \text{82.1\%} & 56.7 & \textbf{\text{74.8\%}}\\
    Dynasor & 54.3 & 9912 & \text{13.4\%} & 86.7 & 6233 & \text{16.3\%} & 90.1 & 3877 & \text{19.2\%} & 50.5 & 4398 & \text{46.3\%} & 70.4 & \text{23.8\%}\\
    DEER & 50.0 & 6873 & \text{40.0\%} & 80.0 & 3231 & \text{56.6\%} & 87.0 & 1854 & \text{61.3\%} & 54.9 & 7033  & \text{14.2\%}  & 68.0 & \text{43.0\%} \\
    CGRS (\textbf{ours}) & 56.7 & 7893 & \text{31.1\%} & 86.7 & 4351 & \text{41.6\%} & 91.3 & 2704 & \text{43.6\%} & 55.2 & 5229 & \text{36.2\%} & \textbf{72.5} & \text{38.1\%} \\
    \midrule
    \multicolumn{15}{c}{\textit{\textbf{Qwen3-8B}}} \\
    Vanilla & 61.1 & 11924 & \text{-} & 89.4 & 7876 &  \text{-} & 92.9 & 5104 & \text{-} & 57.7 & 9104 & \text{-} & 75.3 & \text{-} \\
    TALE & 68.9 & 10942 & \text{8.2\%} & 88.3 & 6872 & \text{12.7\%} & 92.3 & 3885 & \text{23.9\%} & 59.1 & 7113 & \text{21.9\%} & \textbf{77.2} & \text{16.7\%}\\
    NoThinking & 30.0 & 5967 & \text{50.0\%} & 72.5 & 2426 & \text{69.2\%} & 87.1 & 1239 & \text{75.7\%} & 54.2 & 1546 & \text{83.0\%}  & 61.0 & \textbf{\text{69.5\%}}\\
    Dynasor & 62.2 & 10174 & \text{14.7\%} & 89.2 & 6457 & \text{18.0\%} & 91.7 & 3841 & \text{24.7\%} & 57.7 & 5965 & \text{34.5\%}  & 75.2 & \text{23.0\%}\\
    DEER & 45.6 & 7443 & \text{37.6\%} & 79.2 & 3715 & \text{52.8\%} & 88.7 & 1935 & \text{62.1\%} & 59.3 & 7837 &  \text{13.9\%} & 68.2 & \text{41.6\%}\\
    CGRS (\textbf{ours}) & 61.1 & 8792 & \text{26.3\%} & 89.2 & 5595 & \text{29.0\%}  & 93.3 & 3507 & \text{31.3\%} & 59.8 & 6302 &  \text{30.8\%} & 75.9 & \text{29.3\%}\\
    \midrule
    \multicolumn{15}{c}{\textit{\textbf{Qwen3-14B}}} \\
    Vanilla & 68.9 & 11316 & \text{-} & 93.3 & 7190 &  \text{-} & 94.1 & 4551 & \text{-} & 64.0 & 7411 & \text{-} & 80.1 & \text{-}\\
    TALE & 71.1 & 10860 & \text{6.4\%} & 92.5 & 5951 & \text{17.2\%} & 93.7 & 3389 & \text{25.5\%} & 63.8 & 6091 & \text{17.8\%} & 80.3 & \text{16.2\%}\\
    NoThinking & 27.8 & 3689 & \text{67.4\%} & 77.5 & 1616 & \text{77.5\%} & 87.0 & 853 & \text{81.3\%} & 56.9 & 1268 & \text{82.9\%}  & 62.3 & \textbf{\text{77.3\%}}\\
    Dynasor & 65.6 & 9775 & \text{13.6\%} & 90.0 & 6030 & \text{16.1\%} & 84.4 & 3667 & \text{19.4\%} & 64.3 & 5775 & \text{22.1\%}  & 76.1 & \text{17.8\%}\\
    DEER & 56.7 & 6755 & \text{40.3\%} & 90.8 & 4079 & \text{43.3\%} & 91.7 & 1956 & \text{57.0\%} & 58.2 & 6338 & \text{14.5\%} & 74.4 & \text{38.8\%}\\
    CGRS (\textbf{ours}) & 70.0 & 8662 & \text{23.5\%} & 93.3 & 5076 & \text{29.4\%} & 94.5 & 3235 & \text{28.9\%} & 65.2 & 5953 & \text{19.7\%} & \textbf{80.8} & \text{25.4\%}\\
    \midrule
   \multicolumn{15}{c}{\textit{\textbf{Qwen3-32B}}} \\
    Vanilla & 67.8 & 11022 & \text{-} & 95.8 & 6794 & \text{-} & 93.9 & 4473 & \text{-} & 71.4 & 7062 & \text{-} & 82.2 & \text{-}\\
    TALE & 67.8 & 10688 & \text{3.0\%} & 93.3 & 6533 & \text{3.8\%} & 93.6 & 3857 & \text{13.8\%} & 67.0 & 5916 & \text{16.2\%} & \textbf{80.4} & \text{9.2\%}\\
    NoThinking & 41.1 & 5635 & \text{48.9\%} & 75.0 & 2221 & \text{67.3\%} & 87.0 & 1054 & \text{76.4\%} & 56.6 & 917 & \text{87.0\%} & 64.9 & \textbf{69.9\%}\\
    Dynasor & 64.4 & 9518 & \text{13.6\%} & 92.5 & 5521 & \text{18.7\%} & 85.2 & 3486 & \text{22.1\%} & 58.1 & 3161 & \text{55.2\%} & 75.0 & \text{27.4\%}\\
    % NoWait &  &  &  & 91.7 & 4859 &  &  &  &  & 56.1 & 4455 &  &  & \text{-\%}\
    DEER & 68.9 & 8697 & \text{21.1\%} & 88.2 & 4878 & \text{28.2\%} & 93.2 & 2533 & \text{43.4\%} & 69.6 & 6310 & \text{10.6\%} & 80.0 & \text{25.8\%}\\
    CGRS (\textbf{ours}) & 65.6 & 8128 & \text{26.3\%} & 94.2 & 4766 & \text{29.8\%} & 93.1 & 2993 & \text{33.1\%} & 64.0 & 4535 & \text{35.8\%} & 79.2 & \text{31.2\%}\\
    \midrule
    \multicolumn{15}{c}{\textit{\textbf{DeepSeek-R1-Distill-Qwen-7B}}} \\
    Vanilla & 52.2 & 10662 & \text{-} & 87.5 & 5861 &  \text{-} & 91.3 & 3787 & \text{-} & 36.2 & 7191 & \text{-} & 66.8 & \text{-} \\ 
    TALE & 48.9 & 9727 & \text{8.8\%} & 86.7 & 5107 & \text{19.9\%} & 89.1 & 2657 & \text{29.8\%} & 36.2 & 4938 & \text{31.3\%} & \textbf{65.2} & \text{20.7\%}\\
    NoThinking & 32.2 & 6680 & \text{37.3\%} & 75.8 & 2499 & \text{57.4\%} & 80.9 & 1173 & \text{65.4\%} & 37.9 & 1312 & \text{81.8\%} & 56.7 & \textbf{\text{61.4\%}}\\
    Dynasor & 47.8 & 8334 & \text{21.8\%} & 84.2 & 5201 & \text{11.3\%} & 81.8 & 2070 & \text{45.3\%} & 22.2 & 561 & \text{92.2\%} & 59.0 & \text{42.7\%} \\
    DEER & 47.8 & 9288 & \text{12.9\%} & 88.3 & 4670 & \text{20.3\%} & 89.6 & 2272 & \text{40.0\%} & 33.1 & 6457 & \text{10.2\%} & 64.7 & \text{20.9\%} \\
    CGRS (\textbf{ours}) & 52.2 & 7597 & \text{28.7\%} & 88.3 & 3406 & \text{41.9\%} & 87.6 & 1867 & \text{50.7\%} &  32.8 & 3876 & \text{46.1\%} & \textbf{65.2} & \text{41.9\%} \\
    \midrule
    \multicolumn{15}{c}{\textit{\textbf{DeepSeek-R1-Distill-Llama-8B}}} \\
    Vanilla & 37.7 & 11898 & \text{-} & 84.2 & 6374 &  \text{-} & 85.7 & 4087 & \text{-} &  39.2  & 8096 & \text{-} & 61.7 & \text{-} \\
    TALE & 40.0 & 11141 & \text{6.4\%} & 85.0 & 5915 & \text{7.2\%} & 84.0 & 3541 & \text{13.4\%} & 46.1 & 5573 & \text{31.2\%} & 63.8 & \text{14.5\%}\\
    NoThinking & 40.0 & 11242 & \text{5.5\%} & 82.5 & 5796 & \text{9.1\%} & 83.3 & 2405 & \text{41.2\%} & 36.2 & 6638 & \text{18.0\%} & 60.5 & \text{18.4\%}\\
    Dynasor & 37.7 & 10368 & \text{12.9\%} & 84.2 & 5681 & \text{10.9\%} & 85.0 & 3585 & \text{12.3\%} & 31.8 & 3095 & \text{61.8\%} & 59.7 & \text{24.4\%}\\
    DEER & 42.2 & 9778 & \text{17.8\%} & 80.8 & 5480 & \text{14.0\%} & 82.3 & 2722 & \text{33.4\%} & 41.8 & 7434 & \text{8.2\%} & 56.4 & \textbf{\text{35.7\%}} \\
    CGRS (\textbf{ours}) & 47.8 & 9536 & \text{19.9\%} & 86.7 & 4899 & \text{23.1\%} & 84.7 & 3254 & \text{20.4\%} & 39.6 & 7221 & \text{10.8\%}  & \textbf{64.7} & \text{18.5\%} \\
    \midrule
    \multicolumn{15}{c}{\textit{\textbf{QwQ-32B}}} \\
    Vanilla & 71.1 & 11026 & \text{-} & 89.1 & 7210 &  \text{-} & 94.2 & 4216 & \text{-} & 66.4  & 7269 & \text{-} & 80.2 & \text{-}  \\
    TALE & 61.1 & 10888 & \text{1.3\%} & 90.8 & 6522 & \text{9.5\%} & 94.0 & 3533 & \text{16.2\%} & 65.5 & 6482 & \text{10.8\%} & 77.8 & \text{9.5\%}\\
    NoThinking & 62.2 & 11688 & \text{-6.0\%} & 88.3 & 7493 & \text{-3.9\%} & 94.2 & 4276 & \text{-1.4\%} & 65.2 & 7604 & \text{-4.6\%} & 77.5 & \text{-4.0\%}\\
    Dynasor & 64.4 & 9733 & \text{11.7\%} & 90.0 & 7185 & \text{0.3\%} & 94.0 & 4156 & \text{1.4\%} & 41.9 & 2478  & \text{65.9\%} & 72.6 & \text{19.9\%}\\
    DEER & 65.6 & 10015 & \text{9.2\%} & 92.5 & 6324 & \text{12.3\%} & 94.1 & 3359 & \text{20.3\%} & 66.5 & 6453 & \text{11.2\%} & 79.7 & \text{13.3\%} \\
    CGRS (\textbf{ours}) & 68.9 & 8202 & \text{25.6\%} & 93.3 & 4771 & \text{33.8\%} & 94.2 & 2810 & \text{33.3\%} & 67.0 & 5141 & \text{29.3\%} & \textbf{80.8} & \textbf{\text{30.5\%}}  \\
    \bottomrule
    \end{tabular}}
    \caption{Comparison across models of different scales and multiple methods. Each experiment is repeated three times and the average results are reported. ``Acc" (\%) and ``Len" (in tokens) denote the accuracy and response length, respectively. ``LR" is the average length reduction ratio relative to Vanilla. $\uparrow$ ($\downarrow$) indicates that the higher (lower) the result, the better the performance. The best results (except for Vanilla) are highlighted in \textbf{bold}.}
    \label{tab:main_results}
\end{table*}

In this section, we evaluate the proposed CGRS method on multiple benchmark datasets and models to demonstrate its effectiveness in enabling efficient reasoning.

\subsection{Experimental Setup} \label{sec:exp_setup}

\paragraph{Benchmark Datasets.} We evaluate on three mathematical reasoning benchmarks: {AIME24} \cite{aime} that contains $30$ high-difficulty problems from the 2024 {American Invitational Mathematics Examination}, {AMC23} \cite{amc23} that includes $40$ problems from the 2023 {American Mathematics Competitions}, and {MATH500} \cite{math500} that has $500$ multi-step problems covering algebra, geometry, and probability, curated by OpenAI, and one scientific reasoning benchmark: GPQA Diamond (abbreviated as GPQA-D) \cite{gpqa}, a dataset of $198$ multiple-choice questions on biology, chemistry, and physics, at the post-graduate level.

\paragraph{Models.} We conduct experiments on the a series of open-source models spanning different architectures and scales, including \textit{DeepSeek-R1-Distill} series of models (Qwen-7B and Llama-8B)~\citep{guo2025deepseekr1}, \textit{QwQ-32B}~\citep{qwq32b}, and the \textit{Qwen3} family (4B, 8B, 14B, and 32B)~\citep{qwen3}. 

\paragraph{Implementation Details.} All experiments are conducted using the open-source \texttt{vLLM} framework \cite{vllm} to ensure high-throughput and memory-efficient inference. All decoding use temperature $0.6$ and top-p $0.95$. Each experiment is repeated three times and the average results are reported. The reflection suppression threshold $\delta$ in our CGRS method is set to $0.9$.

\paragraph{Baselines.}
The proposed CGRS method is compared with three types of baselines:
(i) \textbf{Vanilla} that performs standard decoding without any intervention,
(ii) prompt-guided methods, including \textbf{NoThinking} \cite{ma2025reasoningmodelseffectivethinking} that prompts the model to bypass intermediate reasoning entirely and generate the final answer directly, and \textbf{TALE}~\cite{han-etal-2025-token} that prompts the model to solve the problem within token budgets (in our experiments,
we set this budget based on the actual token length generated by CGRS), and
(iii) decoding-manipulation methods, including \textbf{Dynasor} \cite{fu2025efficientlyscalingllmreasoning} that periodically requests intermediate answers at fixed token intervals and exits early if multiple consecutive answers match and \textbf{DEER} \cite{yang2025dynamicearlyexitreasoning} dynamically truncates chain-of-thought generation by detecting high-confidence intermediate answers at transition cues such as ``Wait".

\paragraph{Evaluation Metrics.} We evaluate performance using three metrics: (i) pass@1 accuracy (\textbf{Acc}) that is the proportion of problems correctly solved on the first attempt, (ii) average output token length (\textbf{Len}) that serves as a proxy for reasoning cost during inference, and (iii) the average length reduction ratio (\textbf{LR}) that measures the percentage decrease in output token length compared to the Vanilla method, with higher values indicating higher compression.

\subsection{Comparison with State-of-the-art Methods}

In Table \ref{tab:main_results}, we compare the proposed CGRS method with previous efficient reasoning methods in terms of accuracy and output length. The evaluation covers diverse model scales and architectures, including \textit{DeepSeek-R1-Distill-Qwen-7B}, \textit{DeepSeek-R1-Distill-Llama-8B}, \textit{QwQ-32B}, and \textit{Qwen3-4/8/14/32B} across four reasoning benchmark datasets (i.e., AIME24, AMC23, MATH500, and GPQA-D). 

As can be seen, CGRS consistently achieves the optimal balance  between average length reduction ratio and average accuracy preservation across diverse models, demonstrating its effectiveness for efficient reasoning. For example, on the \textit{Qwen3-14B} model, CGRS reduces output length by $25.4\%$ while maintaining accuracy. In contrast, while baseline methods achieve greater length reduction ($77.3\%$ of NoThinking and $38.8\%$ of DEER), they suffer significant accuracy degradation ($17.8\%$ and $5.7\%$ drops respectively). TALE maintains comparable accuracy but achieves less length reduction than CGRS ($16.2\%$ vs. $25.4\%$).

Moreover, the proposed CGRS method demonstrates reliable effectiveness in every test case, consistently achieving $18.5\%$ to $41.9\%$ length reduction with negligible accuracy drop (up to $3\%$), while the baseline methods often perform unsatisfactorily. 
For example, NoThinking prompts models to bypass slow thinking processes, which always yields high compression rates but severely degrades accuracy (e.g., $65.4\%$ length reduction and $9.4\%$ accuracy drop on MATH500 with \textit{DeepSeek-R1-Distill-Qwen-7B}). 
Early-exit approaches like Dynasor and DEER also sometimes underperform, e.g., on AMC23 with \textit{DeepSeek-R1-Distill-Llama-8B}, they only achieve $10.9\%$ and $14.0\%$ compression, respectively, with DEER additionally reducing accuracy. Remarkably, in the same test case, CGRS attains $23.1\%$ length reduction while slightly improving accuracy. 

Notably, on the \textit{QwQ-32B} model, the baseline methods (TALE, NoThinking, Dynasor, and DEER) achieve only minor reduction ratios across all four benchmark datasets. 
This limitation occurs because these methods rely on hard-coded reflection boundaries that assume reasoning terminates at the \texttt{</think>} token, while \textit{QwQ-32B} often continues reflection behaviors beyond this token \cite{yang2025dynamicearlyexitreasoning}. 
In contrast, the proposed CGRS method dynamically regulates reflection triggers based on confidence scores, which is independent on the end-of-think token, enabling CGRS to maintain strong compression rates ($30.5\%$) while preserving accuracy in this case. 

In summary, these results demonstrate the superior performance of the proposed CGRS method over all baseline approaches across various models and datasets, indicating its effectiveness for efficient reasoning in LRLMs.

\subsection{Effectiveness of Reflection Suppression} 

As introduced in Section~\ref{sec:method}, CGRS proactively suppresses redundant reflection behaviors to alleviate the overthinking problem. To evaluate the effectiveness of this mechanism, in this section, we analyze two key metrics: (i) the frequency of reflection triggers (specifically the words ``Wait", ``But", ``Alternatively", and ``Hmm"), and (ii) the answer length distribution, comparing CGRS against the Vanilla baseline. The experiments employ the \textit{DeepSeek-R1-Distill-Qwen-7B} model on the AIME24 benchmark, following the same experimental setup described in Section~\ref{sec:exp_setup}.

\begin{figure}[!t]
\centering
\includegraphics[width=0.48\textwidth]{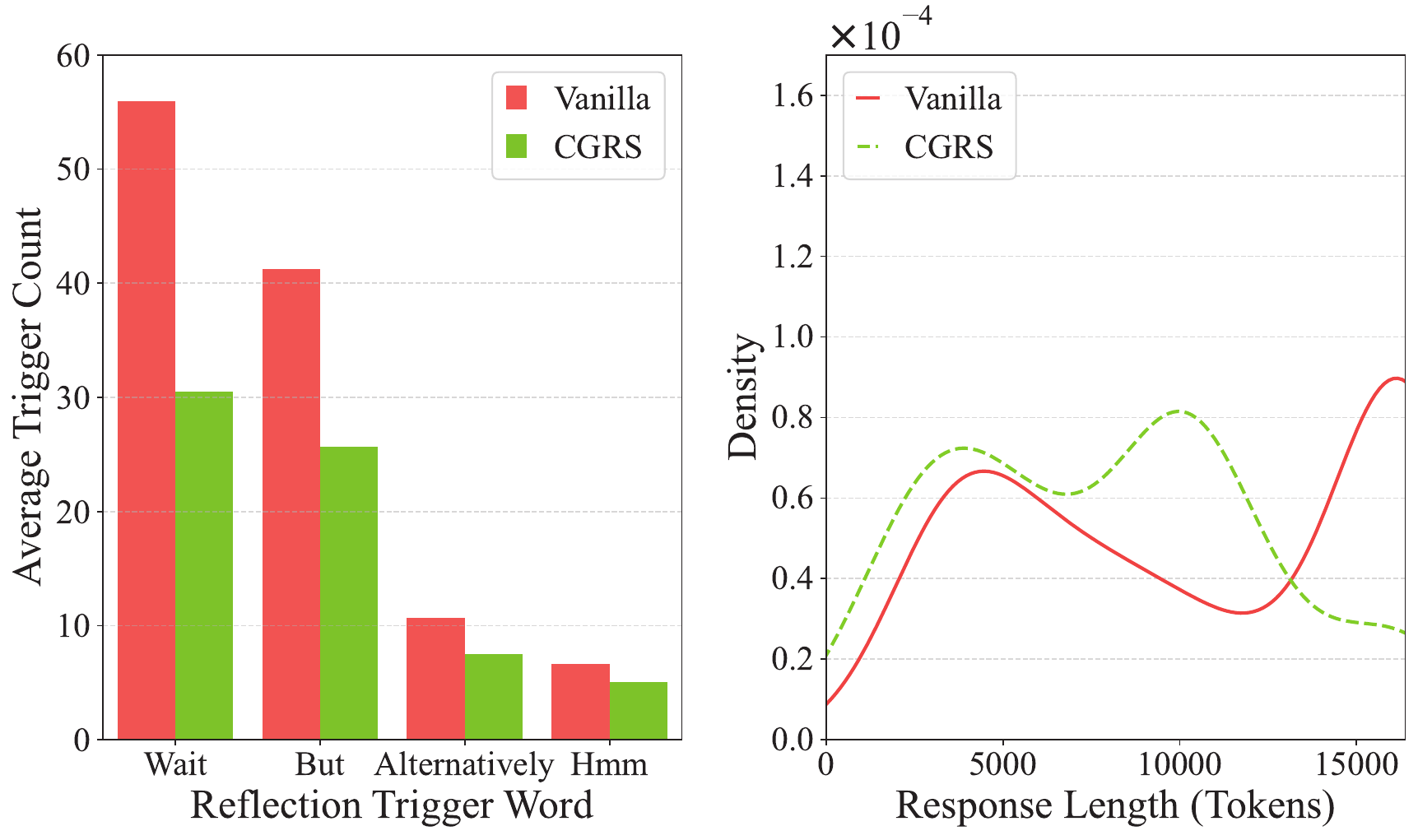}
    \caption{Comparison of Vanilla and CGRS methods in terms of the frequency of reflection triggers (\textbf{left}) and the answer length distribution (\textbf{right}).}
    \label{fig:exp2}
\end{figure}

The results are presented in Figure~\ref{fig:exp2}. Compared to the Vanilla baseline, CGRS demonstrates: (i) a significant reduction in reflection trigger words (particularly ``Wait" and ``But"), and (ii) a more concentrated and significantly shorter length distribution. These findings clearly indicate effective suppression of reflection behaviors. When combined with the findings in Table~\ref{tab:main_results} showing that CGRS achieves 28.7\% length reduction without accuracy decrease on AIME24 using the \textit{DeepSeek-R1-Distill-Qwen-7B} model, these results demonstrate that CGRS effectively alleviate the overthinking problem and enable efficient reasoning.

\subsection{Effectiveness of Certainty-Guided Probability}

As described in Section~\ref{sec:method}, CGRS dynamically suppresses the LRLM's tendency to generate reflection triggers based on the certainty of its current response. In this section, we evaluate the effectiveness of this certainty-guided suppression probability (Eq. (\ref{eq:suppression_prob})) by comparing it against fixed suppression probabilities: $p=0$ (the Vanilla baseline described in Section \ref{sec:exp_setup}), $0.25$, $0.5$, and 
$1$ (where no reflection triggers are sampled during inference). We conduct experiments on the AMC23 dataset using the \textit{DeepSeek-R1-Distill-Qwen-7B} model, following the same setup as in Section~\ref{sec:exp_setup}.

\begin{table}[!t]
\centering
\begin{tabular}{lcc}
\toprule
 & \textbf{Acc}$\uparrow$ & \textbf{Len}$\downarrow$ \\
\midrule
$p=0$ (the Vanilla baseline) & 87.5 & 5861 \\
$p=0.25$ & 81.7 & 3729 \\
$p=0.5$ & 80.0 & 3266 \\
$p=1.0$ & 76.7 & 2373 \\
Certainty-Guided (Eq. (\ref{eq:suppression_prob})) & 88.3 & 3406 \\
\bottomrule
\end{tabular}
\caption{Comparison of certainty-guided suppression probability (Eq. (\ref{eq:suppression_prob})) with fixed probabilities.}
\label{tab:ablation_confidence}
\end{table}

\begin{figure*}[!t]
 \centering
 \includegraphics[width=1.0\textwidth]{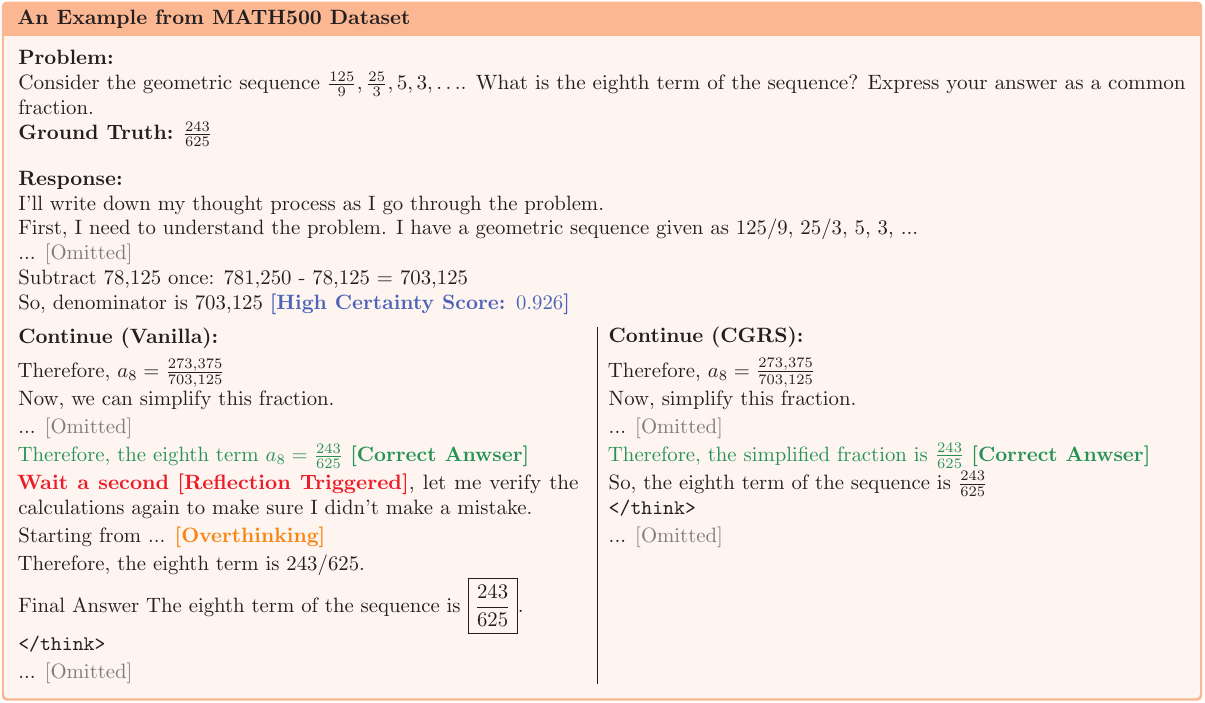}
 \caption{Comparison of the responses from Vanilla and the proposed CGRS (with $\delta$ = 0.9) methods on an example from the MATH500 dataset using the \textit{DeepSeek-R1-Distill-Qwen-7B} model.
 }
 \label{fig:case_study}
\end{figure*}

The results are shown in Table~\ref{tab:ablation_confidence}. As can be seen, as $p$ increases, the response length decreases significantly while the accuracy also drops considerably. This indicates that important reflection behaviors are being suppressed in the slow thinking process. In contrast, our certainty-guided method achieves a 41.9\% reduction in length without compromising accuracy, demonstrating that this certainty-based control effectively suppresses unproductive reflections while preserving necessary ones.

\subsection{Case Study}
In Figure~\ref{fig:case_study}, we present a representative case from the MATH 500 dataset, solved by the \textit{DeepSeek-R1-Distill-7B} model, to demonstrate CGRS's mechanism and advantages. 

During the initial reasoning phase, the model executes routine computations. 
Since the certainty scores for these intermediate steps remain below a predefined threshold, CGRS remains inactive, permitting the model to freely explore and revise its reasoning path—similar to standard autoregressive inference.

The turning point occurs when a high certainty score of $0.926$ is detected, triggering the CGRS mechanism. In the Vanilla baseline, the model performs redundant reflection even after reaching a correct intermediate answer—re-examining previous steps (e.g., common ratio, exponentiation, and multiplication in this case). This unnecessary verification increases computational overhead without enhancing accuracy.
In contrast, CGRS promotes efficient and goal-directed behavior. It suppresses reflection-inducing tokens, preventing re-evaluation of outputs it considers reliable. Instead, the model proceeds directly to the next logical operation (e.g., fraction simplification). This targeted intervention ensures high-confidence reasoning proceeds without unnecessary detours, significantly reducing both token count and inference time.

This case study offers qualitative evidence that CGRS effectively alleviate overthinking by utilizing internal certainty estimates to suppress redundant reasoning steps.

% ============= 4 EXPERIMENT ================

% ============= 5 CONCLUSION ================
\section{Conslusion}

In this paper, we propose {Certainty-Guided Reflection Suppression (CGRS)}, a novel approach to address the overthinking problem in Large Reasoning Language Models (LRLMs) and enable efficient reasoning. CGRS is a lightweight, certainty-guided decoding strategy that dynamically suppresses reflection triggers when the model exhibits high confidence in its current reasoning trajectory. Our method is model-agnostic, requires no retraining or architectural modifications, and can be seamlessly integrated into existing autoregressive generation pipelines. Extensive experiments on open-source LRLMs across four reasoning benchmarks demonstrate that CGRS achieves token usage reductions of up to $41.9\%$ while preserving answer accuracy. These results establish that reflective behaviors in LRLMs can be effectively modulated during inference through certainty-aware decoding. Our work advances the development of efficient reasoning systems by demonstrating how model behavior can be aligned with internal certainty signals, providing a practical pathway toward more scalable LLM deployments.

\section*{Acknowledgments}

This work is supported in part by the State Key Laboratory of General Artificial Intelligence. DH is supported by National Science Foundation of China (NSFC62376007) and Beijing Natural Science Foundation (Z250001).

% ============= 5 CONCLUSION ================

\bibliography{aaai2026}
\clearpage
\appendix
\section{Additional Details of Reflection Triggers Selection in Section 3.3}

As described in Section 3.3, we focus on four primary trigger words (``Wait'', ``But'', ``Alternatively'', and ``Hmm'') in our experiments. For each trigger word, we identify all natural language variants present in the tokenizer's vocabulary. For example, the keyword ``Wait" has variants such as ``{.Wait,}" ``{\textvisiblespace WaitForSeconds}," ``{.wait}," ``{.WaitFor}," ``{ĉwait},". Then we analyze their generation frequencies in reasoning, in order to filter out tokens that are unlikely to appear (e.g., ``{.Wait}") during actual reasoning. This analysis uses reasoning traces from the \textit{DeepSeek-R1-Distill-Qwen7B} model on both the AIME24 and AMC23 benchmarks. We examine two tokenizer types: \texttt{LlamaTokenizerFast} (for \textit{DeepSeek-R1-Distill-Llama-8B}) and \texttt{Qwen2Tokenizer} (for \textit{DeepSeek-R1-Distill-Qwen-7B}, the \textit{Qwen-3} family, and \textit{QwQ-32B}). The complete list of filtered trigger words (including variants) and their corresponding token IDs are provided in Table~\ref{tab:triggers}.

\section{Examples of Redundant Reflection Behaviors}

In this section, we present qualitative examples demonstrating redundant reflection behaviors in LRLM reasoning. We analyze responses from the \textit{DeepSeek-R1-Distill-Qwen-7B} model on a complex combinatorial problem in the AIME24 dataset.

As shown in Table~\ref{tab:redundant_reflection}, the model initially identifies a correct or promising approach but subsequently enters a prolonged reflective phase. This phase is characterized by ineffective self-correction, repeated attempts at previously explored approaches, and overthinking trivial details—often initiated by reflection triggers (like ``Wait'', ``But'' and ``Alternatively''). Despite the extended reasoning, these additional steps fail to improve answer quality and, in some cases, introduce new errors or overwrite correct logic. These failure cases underscore the need for dynamic reflection control during inference. By identifying the onset of these inefficient reflective loops, CGRS helps maintain concise and effective reasoning without suppressing productive analysis.

% \begin{table}[h]
% \centering
% \caption{Core Hesitation and Transition Tokens (Qwen Model)}
% \begin{tabular}{ll}
% \toprule
% \textbf{Keyword Form} & \textbf{Token ID} \\
% \midrule
% Wait & 14190 \\
% \textvisiblespace Wait & 13824 \\
% But & 3983 \\
% \textvisiblespace But & 1988 \\
% wait & 11489 \\
% \textvisiblespace wait & 3783 \\
% but & 8088 \\
% \textvisiblespace but & 714 \\
% Alternatively & 38478 \\
% Alternative & 41109 \\
% Alternative & 75763 \\
% Alternatively & 92014 \\
% Hmm & 80022 \\
% \textvisiblespace Hmm & 88190 \\
% \bottomrule
% \end{tabular}
% \label{tab:qwen-triggers}
% \end{table}

% \begin{table}[h]
% \centering
% \caption{Core Hesitation and Transition Tokens (Llama Model)}
% \begin{tabular}{ll}
% \toprule
% \textbf{Keyword Form} & \textbf{Token ID} \\
% \midrule
% Wait & 14524 \\
% \textvisiblespace Wait & 14144 \\
% \textvisiblespace But & 2030 \\
% \textvisiblespace but & 719 \\
% But & 4071 \\
% \textvisiblespace wait & 3868 \\
% wait & 11748 \\
% \textvisiblespace Alternatively & 39578 \\
% \textvisiblespace Alternative & 42209 \\
% Alternative & 76863 \\
% Alternatively & 93114 \\
% Hmm & 81122 \\
% \textvisiblespace Hmm & 89290 \\
% \bottomrule
% \end{tabular}
% \label{tab:llama-triggers}
% \end{table}

\begin{table}[!t]
% \centering
\begin{tabular}{lcc}
\toprule
\textbf{Trigger Words} & \textbf{Qwen Token ID} & \textbf{Llama Token ID} \\
\midrule
Wait & 14190 & 14524 \\
\textvisiblespace Wait & 13824 & 14144 \\
wait & 11489 & 11748 \\
\textvisiblespace wait & 3783 & 3868 \\
\midrule
But & 3983 & 4071 \\
\textvisiblespace But & 1988 & 2030 \\
but & 8088 & 8248 \\
\textvisiblespace but & 714 & 719 \\
\midrule
Alternatively & 38478 & 93114 \\
\textvisiblespace Alternatively & 38478 & 39578 \\
Alternative & 75763 & 76863 \\
\textvisiblespace Alternative & 41109 & 42209 \\
\midrule
Hmm & 80022 & 81122 \\
\textvisiblespace Hmm & 88190 & 89290 \\
\bottomrule
\end{tabular}
\caption{Reflection trigger words and their corresponding token IDs in \texttt{Qwen2Tokenizer} and \texttt{LlamaTokenizerFast} tokenizers. \textvisiblespace~denotes a space character.}
\label{tab:triggers}
\end{table}

\section{Ablation on Threshold $\delta$}

The results are shown in Table~\ref{tab:ablation_delta}. As $\delta$ decreases, the model becomes less strict in suppressing uncertain reasoning steps, leading to longer responses but lower accuracy. Specifically, when $\delta$ drops from 0.9 to 0.1, accuracy decreases from 88.3\% to 72.5\%, while the length reduction (LR) weakens from 41.9\% to 53.7\%. This demonstrates the trade-off controlled by $\delta$: higher thresholds enforce more confident reasoning and concise outputs, whereas lower thresholds allow more exploratory but error-prone reasoning steps.

\begin{table}[!t]
\centering
\begin{tabular}{lcc}
\toprule
\textbf{$\delta$} & \textbf{Accuracy}$\uparrow$ & \textbf{LR}$\downarrow$ \\
\midrule
0.9 & 88.3\% & 41.9 \\
0.5 & 76.2\% & 48.5 \\
0.1 & 72.5\% & 53.7 \\
\bottomrule
\end{tabular}
\caption{Effect of threshold $\delta$ on accuracy and length reduction (LR) on AMC23 and R1-Distill-Qwen-7B.}
\label{tab:ablation_delta}
\end{table}

\begin{table*}[t]
    \centering 
    \begin{tabular}{p{0.1\textwidth} p{0.5\textwidth} p{0.3\textwidth}} 
        \toprule
         & \textbf{Reflection Examples} & \textbf{Analysis} \\
        \midrule
        \textbf{Case 1: Failed self-correction.} & ``Therefore, $r(S)$ is $S'$ which is equivalent to $S \cap r^{-1}(S') = S$. \textbf{[Correct]} \textit{Wait}, maybe that's not the right way. \textit{Alternatively}, since $r$ is a permutation, $r(S)$ is just another set of the same size as $S$. So, if $S'$ is a subset of $S$, then $r(S)$ must be a subset of $S'$ which would require that $S'$ is a subset of $r^{-1}(S')$. \textit{But} since $r$ is a rotation, it's a bijection, so $r^{-1}(S')$ is just another set of size $|S|$. \textbf{[Wrong]}" & \textbf{correct to wrong:} The initial thought process was on the right track, but the self-correction led to incorrect reasoning steps. \\
        \addlinespace
        & ``\textit{Wait}, but perhaps the number is 105. \textbf{[Wrong]} \textit{Wait}, but 105 is $2 \cdot 10^2$, which is 105. \textit{Wait}, perhaps the number is 105. \textbf{[Wrong]}" & \textbf{wrong to wrong:} Repeated incorrect self-correction, circling back to the same incorrect value. \\
        \addlinespace % Added addlinespace for consistent spacing
        \midrule % Replaced \newpage with \midrule for proper table separation
        \textbf{Case 2: Reattempt a previously tried approach.} & 
        %``\textit{Wait}, so $\{1,2,4\}$ is a three-element set that cannot be rotated to lie entirely within $S'$, which is five-element set. Therefore, not all three-element sets satisfy the condition. \textbf{Therefore, I need a better approach.} Perhaps, instead of considering all subsets, I should model this as the number of subsets $S$ such that there exists a rotation $r$ where $r(S)$ is disjoint from $S$. \textbf{[This method was previously tried.]}"
        ``\textit{Wait}, but in this case, it's more specific: the set must be entirely disjoint from some rotated copy of itself. \textit{Hmm}. \textit{Wait}, maybe it's useful to think in terms of necklace structures. A necklace is an equivalence class under rotation, but I'm not sure. \textbf{[The method ``necklace counting" was tried before and failed.]}"
        & This highlights an instance where the model redundantly reapplied a previously attempted strategy it had already recognized as ineffective for the current problem. \\
        \addlinespace % Added addlinespace for consistent spacing
        \midrule
        \textbf{Case 3: Overthinking trivial details.} & ``\textit{Wait}, but actually, for the identity rotation, $r(S) = S$, so $S \cap r(S) = S$, which is not empty unless $S$ is empty. So, for the identity rotation, the condition $S \cap r(S) = \emptyset$ is only satisfied when $S$ is empty. So, for non-empty $S$, the identity rotation cannot satisfy the condition. Therefore, for non-empty $S$, we need some non-identity rotation $r$ where $S \cap r(S) = \emptyset$. \textit{Wait}, actually, the problem says `can be rotated', so it includes the identity rotation. \textit{But} as we saw, for non-empty $S$, the identity rotation only works if $S$ is empty, which is \textbf{trivial}. So, perhaps the problem is really about non-empty $S$ where some non-identity rotation makes $S \cap r(S) = \emptyset$." & This lengthy deliberation focuses on the trivial case of the identity rotation and its implications. This demonstrates a scenario where the model's reflection-driven over-complication of a simple problem result in unnecessarily verbose reasoning. \\
        \bottomrule
    \end{tabular}
    \caption{Examples of redundant reflection behaviors extracted from the response generated by the \textit{DeepSeek-R1-Distill-Qwen-7B} model on a specific problem in AIME24 dataset. The original problem is ``\textit{Each vertex of a regular octagon is independently colored either red or blue with equal probability. The probability that the octagon can then be rotated so that all of the blue vertices end up at positions where there were originally red vertices is $\tfrac{m}{n}$, where $m$ and $n$ are relatively prime positive integers. What is $m+n$?}''}
    \label{tab:redundant_reflection}
\end{table*}

\section{Relationship Between Certainty and Correctness}

\begin{figure*}[!t]
\centering
\includegraphics[width=0.98\textwidth]{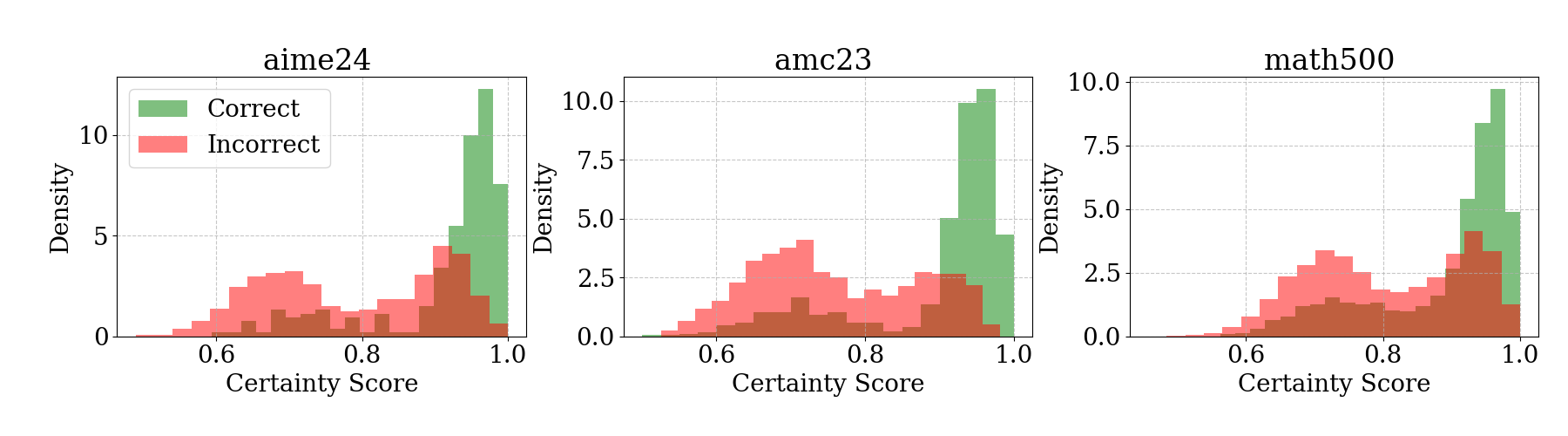}
    \caption{Certainty distributions of probes with correct and incorrect results on AIME24, AMC23, and MATH500.}
    \label{fig:certainty_correctness}
\end{figure*}

Our further analysis correlation between reasoning certainty and correctness. Specifically, we collect reasoning trajectories generated by \texttt{DeepSeek-R1-Distill-Qwen-7B} on three benchmarks: AIME24, AMC23, and MATH500. 
For each trajectory, we insert fixed-interval probing points every 128 tokens to examine 
(1) the correctness of the probed content, and 
(2) the certainty score computed by our method. 
As shown in Figure~\ref{fig:certainty_correctness}, reasoning steps with high certainty 
(e.g., $>0.9$) are significantly more likely to be correct, 
indicating that the model’s self-estimated certainty serves as a reliable indicator of reasoning quality.

\end{document}